\documentclass[letterpaper]{article} 
\usepackage[preprint]{aaai2027}  
\usepackage[hyphens]{url}  
\usepackage{graphicx} 
\usepackage{natbib}  
\usepackage{caption} 
\usepackage{algorithm}
\usepackage{algorithmic}

\usepackage{newfloat}
\usepackage{listings}
\DeclareCaptionStyle{ruled}{labelfont=normalfont,labelsep=colon,strut=off} 
\floatstyle{ruled}
\newfloat{listing}{tb}{lst}{}
\floatname{listing}{Listing}

\usepackage{amsmath}
\usepackage{amssymb}
\usepackage{subfigure}
\usepackage{multirow}
\usepackage{booktabs}

\title{The Grokked Illusion: True Equilibrium Mitigates Catastrophic Forgetting}

\title{The Grokked Illusion: True Equilibrium Mitigates Catastrophic Forgetting}
\author {
  Xiaotian Zhang\textsuperscript{\rm 1},
  Lai Shun Chan\textsuperscript{\rm 1},
  Yue Shang\textsuperscript{\rm 1,\rm 2},
  Ge Zhang\textsuperscript{\rm 1}\corresponding,
  Entao Yang\textsuperscript{\rm 3}\corresponding\\ 
}
\affiliations {
    \textsuperscript{\rm 1}Department of Physics, City University of Hong Kong, Hong Kong, China\\
    \textsuperscript{\rm 2}Department of Physics and Astronomy, University of Pennsylvania, Philadelphia, PA, USA\\
    \textsuperscript{\rm 3}Innovation Campus Delaware, Air Liquide, Newark, DE, USA\\
    gzhang37@cityu.edu.hk, entao.yang@airliquide.com
}

\begin{document}

\maketitle

\begin{abstract}
While neural networks are typically evaluated by their training and test performance, these metrics do not reveal how robust a learned representation is. 
Recent studies have shown that solutions occupying larger volumes in parameter space, as quantified by Boltzmann entropy, often exhibit superior generalizability compared to those reached by conventional optimization, a phenomenon known as the \textit{high entropy advantage}. 
Here we ask whether this advantage persists beyond generalization. 
Specifically, we investigate models' robustness, the ability to retain the learned knowledge when the model is subsequently trained to acquire new information. 
Using grokking in modular arithmetic as a controlled setting, we design a noise injection experiment to evaluate the robustness difference between AdamW-trained transformers and high-entropy model sampled from Wang-Landau Molecular Dynamics with identical saturated performance. 
By forcing both models to fully remember new data with random labels, we find that AdamW-trained models suffer from catastrophic forgetting, with original task test accuracy dropping from 100\% to below 75\%, whereas the high-entropy models maintain approximately 95\% test accuracy. 
We term this hidden fragility behind apparent generalization the ``grokked illusion.'' 
Through singular value decomposition of the neural network weights, we discover that high-entropy neural networks possess significantly higher effective rank in attention and MLP layers both before and after noise injection, indicating richer feature representations can serve as a buffer against catastrophic forgetting. 
Our findings demonstrate that perfect generalization does not imply equal robustness, offering a new perspective on what makes a trained model robust to interference. 
 
\end{abstract}

\section{Introduction}

Generalization performance has long been a key metric for evaluating machine learning models, reflecting the ability of neural networks (NNs) to learn knowledge and apply it to unknown data. 
Yet a model that generalizes perfectly to a test set may still be fragile: parameter updates caused by memorizing new input data may lead to a catastrophic collapse of its previously acquired knowledge. 
This disconnect between generalization and robustness has been observed across a wide range of settings, from adversarial examples \cite{goodfellow2015explaining} to catastrophic forgetting in continual learning \cite{mccloskey1989catastrophic}, and it raises a fundamental question: what properties of a NN determine not only its generalizability, but also its robustness to interference?

Modern NNs often have a large number of parameters, enabling them to perfectly fit finite training data and even random noise \cite{zhang2017understanding}. 
One mainstream research direction attempts to characterize NN states through the geometric structure of the parameter space. 
A classic conjecture in this direction is that flatter minima in the loss landscape correspond to higher generalizability \cite{hochreiter1997flat}. 
However, due to the high dimensionality of the landscape, local flatness metrics such as the maximum eigenvalue of the Hessian matrix are usually difficult to calculate and cannot express the full picture of the loss landscape\cite{ghorbani2019investigation, dinh2017sharp}. 
Even worse, the sharpness quantities computed from the Hessian can be arbitrarily reshaped by simply rescaling of the weights, without impacting NNs' generalization performance \cite{dinh2017sharp}. 
A recent study introduced the Boltzmann entropy of a NN state as a global measure of the volume in parameter space corresponding to a given training loss and test performance \cite{yang2026high}. 
Their work reveals the high-entropy advantage: high entropy states that occupy a larger volume in the parameter space exhibit superior generalizability compared to conventionally trained solutions. 
This observation leads naturally to a hypothesis about robustness: if a high-entropy solution resides within a larger basin in parameter space, then when its parameters are displaced by external perturbations, the resulting state is more likely to remain within the same high-entropy region, thereby preserving its learned functionality. 
In other words, the same property that confers a generalization advantage may also confer a robustness advantage. 

To test this hypothesis while controlling for the confounding effect of generalization performance itself, we choose modular arithmetic \cite{power2022grokking} as our experimental setting. 
In this task, both conventionally trained Transformer models and equilibrium states (states with the highest entropy under a given training loss) can achieve identical perfect test accuracy \cite{zhang2026grokking}, allowing us to isolate the effect of entropy on robustness. 
We design a noise injection experiment in which a NN that has already achieved perfect generalization in the original task must additionally memorize new noisy data. 
Despite the continued availability of the original training data during noise injection, AdamW-trained NNs exhibit a severe drop in original task test accuracy from 100\% to below 75\% in the most challenging condition. 
In stark contrast, the equilibrium NNs sampled via Wang-Landau Molecular Dynamics (WLMD) maintain approximately 98\% test accuracy on the original task after memorizing the same noisy data.
We term this phenomenon the ``grokked illusion'', where solutions appear to grok (saturated test performance) for the same task can present different levels of robustness. 
Through singular value decomposition (SVD) of the NN weights, we find that equilibrium NNs possess significantly higher effective rank in attention and MLP layers both before and after noise injection, indicating that richer feature representations serve as a buffer against catastrophic forgetting. 

Our main contributions are three-fold:
\begin{enumerate}
\item We identify the ``grokked illusion,'' a phenomenon in which NNs that achieve perfect generalization via conventional training nevertheless suffer from catastrophic forgetting when faced with new noisy inputs.
\item We demonstrate the high-entropy robustness advantage: high-entropy states maintain significantly stronger retention of prior knowledge than conventionally trained NNs. 
\item We reveal through singular value analysis that this robustness advantage arises from higher effective rank, linking the statistical physics of the parameter space to the robustness of learned representations. 
\end{enumerate}

\section{Methodology}

\subsection{Training Protocols}

We adopt the modular arithmetic task $(x + y \bmod 67)$ as our primary benchmark, following the setup of \cite{nanda2023progress, zhang2026grokking}. 
The model is a single-layer Transformer with vocabulary size $d_{\text{vocab}}=68$, embedding dimension $d_{\text{model}}=128$, MLP hidden dimension $d_{\text{mlp}}=512$, $d_{\text{head}}=32$, and $4$ attention heads, amounting to approximately $2.15\times 10^5$ parameters, which is severely overparameterized relative to the training set size of $2244$ samples (approximately half of the $67\times 67$ possible input pairs). 

To obtain high-entropy NNs state, we employ the Wang-Landau Molecular Dynamics (WLMD) method \cite{Wang2001} introduced in prior work \cite{yang2026high, zhang2026grokking}. 
The WLMD algorithm samples the Boltzmann entropy landscape $S(\mathcal{L}_{\text{train}}, \mathcal{A}_{\text{test}}) = \log V(\mathcal{L}_{\text{train}}, \mathcal{A}_{\text{test}})$, where $V$ denotes the volume in the parameter space corresponding to a given training loss $\mathcal{L}_{\text{train}}$ and test accuracy $\mathcal{A}_{\text{test}}$. 
We restrict the sampling to configurations with weight norm $\|w\| = 30$, following the ``Goldilocks zone'' identified in \cite{liu2022towards}, and used the ``entropy bias'' method \cite{zhang2026grokking} to constrain the sampling to the range $\ln(\mathcal{L}_{\text{train}}) \in [-20, 1]$, which ensure that the entropy landscape within the low training loss range of interest converges correctly. 
The simulation is run for $1.2\times 10^8$ WLMD epochs until the histogram flatness criterion is satisfied, indicating convergence of the entropy landscape. 
From the converged equilibrium curve, we select a NN state that achieves test accuracy $=100\%$ with $\ln(\mathcal{L}_{\text{train}}) \approx -1$. 
We refer to such states as \emph{equilibrium states} (or high-entropy states), i.e., configurations that maximize entropy for a given training loss. 

For comparison, we train NNs using the AdamW optimizer with the same weight norm constraint $\|w\| = 30$. 
The weight norm is enforced after each parameter update via linear scaling following \cite{liu2023grokking}: after each AdamW update, we rescale all parameters by a factor such that the resulting NN satisfies $\|w\| = 30$. 
We use a learning rate of $1\times 10^{-2}$, weight decay of $1.0$, and full-batch training for $120$ epochs, producing a NN with $\ln(\mathcal{L}_{\text{train}}) \approx -2$ and test accuracy $100\%$. 
Throughout this paper, we denote these two types of NNs as \emph{WLMD-equilibrium} (WLMD) and \emph{AdamW-goldilocks} (AdamW), respectively. 
Both are trained under identical weight norm constraints and achieve perfect test accuracy, allowing us to isolate the effect of entropy on robustness while controlling for generalization performance and weight scale. 

We also examined the robustness of AdamW trained on the full dataset (including test samples) under the same weight norm constraint ($\|w\|=30$) and standard AdamW-trained NNs, with results presented in the Appendix. 
Full details of the computing infrastructure, software dependencies, and hyperparameter configurations are provided in the Appendix. 

\subsection{Noise Injection Protocol}

In this work, we define \emph{robustness} as the ability of a NN to retain previously learned knowledge when its parameters are subsequently updated to accommodate new information. 
We quantify robustness by measuring the NN's test accuracy on the original task after it has fully memorized the injected noisy data (i.e., achieving $>99.8\%$ training accuracy on the combined dataset). 
To evaluate this property while controlling for the confounding effect of generalization performance, we design a noise injection experiment in which a NN that has already mastered the original modular arithmetic task must additionally memorize new noisy data. 
We first expand the vocabulary dimension of the trained NN from $68$ to $136$ by duplicating the embedding and output projection layers, and randomly initializing the newly added parameters. 
The original input tokens occupy indices $0,\dots,67$, while the new tokens for noisy data occupy indices $68,\dots,104$ (corresponding to the 37 distinct numerical symbols and one function symbol). 
The remaining indices $105,\dots,135$ are left unused as redundant vocabulary dimensions. 
This expansion preserves the NN's existing functionality on the original task while equipping it with the capacity to process new inputs. 

We construct three types of noisy datasets, each containing $500$ training samples with labels mapped to the new token indices:
\begin{itemize}
\item \textbf{Random noise}: inputs are randomly generated; 
\item $\mathbf{x^2 + y \bmod 37}$: inputs follow a structured arithmetic rule with modulus $37$, offering partial structural similarity to the original task;
\item $\mathbf{x + y \bmod 37}$: inputs follow the same additive rule as the original task but with a different modulus, representing the closest structural relationship among the three noise types.
\end{itemize}
These three noise types form a spectrum of structural similarity to the original task, ranging from completely unrelated to progressively more aligned, which allows us to examine how the nature of new knowledge affects the retention of previously learned representations.

The expanded NN is then fine-tuned on a mixed training set comprising the original $2244$ training samples and the $500$ noisy samples. 
We use the AdamW optimizer with a learning rate of $1\times 10^{-4}$, full-batch updates, and a maximum of $5000$ epochs. 
Training is terminated when the model achieves $>99.8\%$ training accuracy on the combined dataset, indicating that the NN has fully memorized both the original and the noisy data. 
In practice, this criterion is reached within $4500$ epochs across all settings. 
We monitor the NN's test accuracy on the original task throughout this fine-tuning process. 
Importantly, the training accuracy on the original task data remains at $100\%$ during fine-tuning, confirming that the NN continues to fit the original training data. 
Nevertheless, as we will show, AdamW-trained NNs still exhibit a significant drop in original-task test accuracy, revealing that maintaining perfect fit to the training data does not guarantee the retention of generalizable knowledge. 

\subsection{Singular Value Decomposition and Effective Rank Analysis}

To investigate the mechanistic basis of the observed robustness differences, we perform singular value decomposition (SVD) on the weight matrices of individual layers before and after noise injection. 
For a weight matrix $W \in \mathbb{R}^{m\times n}$, let its singular values be $\sigma_1 \ge \sigma_2 \ge \dots \ge \sigma_k > 0$, where $k = \min(m,n)$. 
We define the normalized singular value distribution as $\tilde{\sigma}_i = \sigma_i / \sum_{j=1}^k \sigma_j$. 
Following \cite{roy2007effective}, the \emph{effective rank} (ER) is defined as:
\[
\text{ER}(W) = \exp\left(-\sum_{i=1}^k \tilde{\sigma}_i \log \tilde{\sigma}_i\right),
\]
which quantifies the number of significant singular directions and serves as a measure of the representational richness of the layer. 
A higher ER indicates a more uniform singular value spectrum, implying that the layer utilizes a broader set of feature dimensions.

We compute the effective rank for all weight matrices in the NN both before and after noise injection. 
Additionally, we visualize the singular value spectra by plotting the sorted singular value index against the corresponding singular values on a log-log scale, which we term the \emph{Singular Value Spectrum} (SVS) plot. 
This visualization provides a direct comparison of the distributional shape of singular values between equilibrium and AdamW-goldilocks NNs.

To complement the effective rank analysis, we also compute the cosine similarity of the flattened parameter vectors between the NNs before and after noise injection, providing a global measure of parameter displacement. 
The cosine similarity is defined as:
\[
\text{Cos}(\theta_{\text{before}}, \theta_{\text{after}}) = 
\frac{\theta_{\text{before}} \cdot \theta_{\text{after}}}{\|\theta_{\text{before}}\| \|\theta_{\text{after}}\|},
\]
where $\theta$ denotes the flattened concatenation of all trainable parameters.
All SVD analyses are conducted on the weight matrices without bias terms, as biases contribute negligibly to the overall representational capacity in this architecture.

\section{Experimental Results}

We conduct noise injection experiments on four types of NN states: equilibrium state sampled by WLMD (denoted as WLMD-equilibrium NN), AdamW with constrained weight norm $\|w\|=30$ (denoted as AdamW-goldilocks NN), standard AdamW without weight norm constraint (denoted as AdamW-standard NN), and AdamW trained on the full dataset (denoted as AdamW-full NN). 
Our main results focus on the comparison between WLMD-equilibrium NN and AdamW-goldilocks NN, as they achieve both perfect test accuracy and similar training loss under identical weight norm constraints, allowing us to isolate the effect of entropy on robustness. 
Results for AdamW-standard and AdamW-full are provided in the appendix. 

\subsection{Robustness Advantage of WLMD-Equilibrium Neural Networks}

We evaluate the robustness of learned knowledge by measuring the original-task test accuracy during fine-tuning on mixed datasets containing both original training data and new noisy data. 
For each experimental condition, we run noise injection with ten different random seeds in the range [40,50). 

\begin{figure*}[t]
\centering
\subfigure[Training dynamics]{
    \includegraphics[width=0.32\linewidth]{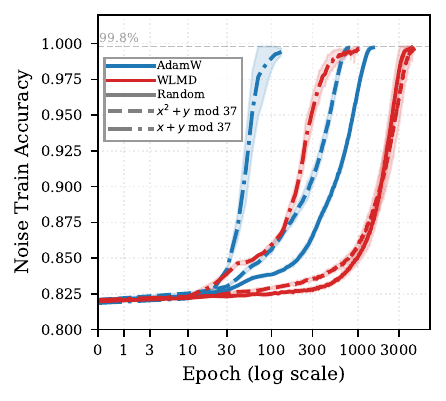}
    \label{fig:epoch-nta}
}
\subfigure[Random noise]{
    \includegraphics[width=0.32\linewidth]{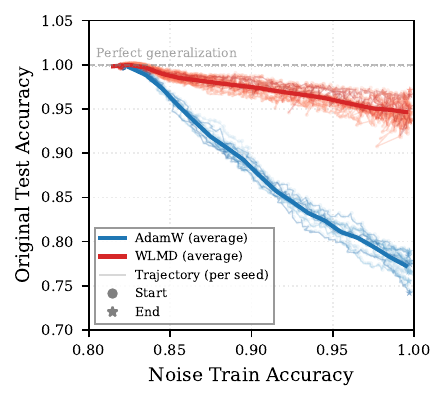}
    \label{fig:trajectory-random}
}
\vspace{0.1cm}

\subfigure[$x^2+y \bmod 37$]{
    \includegraphics[width=0.32\linewidth]{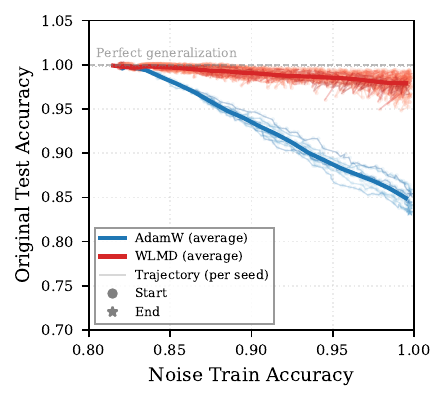}
    \label{fig:trajectory-x2y}
}
\subfigure[$x+y \bmod 37$]{
    \includegraphics[width=0.32\linewidth]{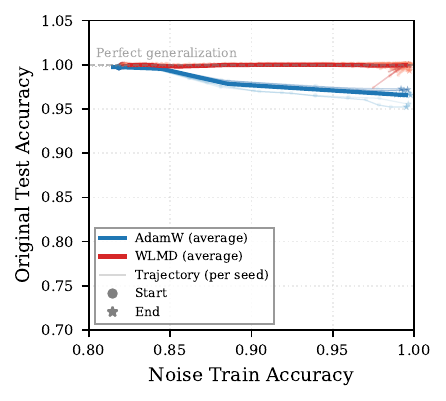}
    \label{fig:trajectory-xy}
}
\caption{Training dynamics and robustness trajectories. (a) Noise training accuracy versus epoch (log scale) for AdamW (blue) and WLMD (red) under all three noise types, with shaded areas indicating $\pm1$ standard deviation across 10 seeds. Solid/dashed/dash-dotted lines correspond to random, $x^2+y \bmod 37$, and $x+y \bmod 37$, respectively. (b--d) Trajectories of (Noise Train Accuracy, Original Test Accuracy) during fine-tuning for random, $x^2+y \bmod 37$, and $x+y \bmod 37$ noise, respectively, with mean trajectories highlighted in blue (AdamW) and red (WLMD). Gray dashed line at $y=1.0$ marks perfect generalization on the original task.}
\label{fig:robustness}
\end{figure*}

Figure~\ref{fig:robustness} presents the training dynamics for three types of noisy data: random noise, $x^2 + y \bmod 37$, and $x + y \bmod 37$. 
Figure~\ref{fig:epoch-nta} reveals that for all three noise types, both WLMD-equilibrium and AdamW-goldilocks NNs can achieve $>99.8\%$ training accuracy on the mixed dataset, confirming that both types of NNs successfully memorize the injected noisy data. 
WLMD-equilibrium NNs consistently require more epochs to memorize the noisy data than AdamW-goldilocks NNs across all noise types, suggesting that NNs at the equilibrium are less sensitive to noise injection and more difficult to adapt. 
Those NNs' behavior on the original task diverges sharply. 
Figures~\ref{fig:trajectory-random}--\ref{fig:trajectory-xy} show the (Noise Train Accuracy, Original Test Accuracy) trajectories for the three noise types. 
In the random noise condition (Fig.~\ref{fig:trajectory-random}), AdamW-goldilocks trajectories exhibit a severe drop from $100\%$ to about $75\%$ in original-task test accuracy. 
This phenomenon, which we term \emph{grokked illusion}, reveals that for the NN generalized at tasks involving grokking through conventional training methods, fitting the training data alone does not guarantee the retention of its knowledge when new irrelevant information is introduced. 
In stark contrast, the WLMD-equilibrium NN maintains approximately $95\%$ test accuracy on the original task after fully memorizing the random noise, demonstrating a substantial robustness advantage. 

For the structured noise types (Figs.~\ref{fig:trajectory-x2y} and~\ref{fig:trajectory-xy}), the robustness gap persists but narrows as the noise becomes more structurally similar to the original task. 
Specifically, for $x^2 + y \bmod 37$, AdamW-goldilocks drops to approximately $84\%$ test accuracy on the original task, while WLMD maintains above $98\%$. 
For $x + y \bmod 37$, where the underlying rule is identical in form to the original task, AdamW-goldilocks achieves approximately $97\%$ and WLMD maintains about $100\%$. 
This monotonic trend indicates that the robustness advantage of WLMD-equilibrium NNs is present across different types of new knowledge, and that the disruption to prior knowledge is inversely related to the structural proximity between the new and original tasks. 

\subsection{Singular Value Analysis and Mechanistic Interpretation}

To better understand the underlying reason for the robustness advantage, we perform singular value decomposition on all weight matrices of the AdamW-goldilocks and WLMD-equilibrium NNs before and after noise injection, with the random noise condition as the primary case study. 
The complete Singular Value Spectrum (SVS) plots for each noise and NN types are provided in the Appendix. 
Table~\ref{tab:svs} reports the effective rank (ER) before and after noise injection, the change in ER, and the cosine similarity (Cos) between the pre- and post-injection weight matrices for each of the nine parameter groups across both NN types. 

\begin{table*}[t]
\centering
\setlength{\tabcolsep}{1.2mm}
\begin{tabular}{lcccccccc}
\toprule
\multirow{2}{*}{Layer} & \multicolumn{4}{c}{AdamW-goldilocks} & \multicolumn{4}{c}{WLMD-equilibrium} \\
\cmidrule(lr){2-5} \cmidrule(lr){6-9}
 & ER$_{\text{pre}}$ & ER$_{\text{post}}$ & $\Delta$ER & Cos & ER$_{\text{pre}}$ & ER$_{\text{post}}$ & $\Delta$ER & Cos \\
\midrule
Embedding       &63.8   &55.3$\pm$0.4   &-8.5$\pm$0.4   &0.9962$\pm$0.0005   &68.7   &64.7$\pm$0.8   &-4.0$\pm$0.8   &0.9892$\pm$0.0018   \\
Pos. Embedding  &1.5    &1.9$\pm$0.0    &+0.4$\pm$0.0    &0.9946$\pm$0.0011   &2.2    &2.2$\pm$0.0    &-0.1$\pm$0.0    &0.9997$\pm$0.0002   \\
$W_Q$           &2.2    &2.4$\pm$0.2    &+0.2$\pm$0.2    &0.9964$\pm$0.0023   &96.8   &68.3$\pm$5.6   &-28.5$\pm$5.6   &0.8383$\pm$0.0379   \\
$W_K$           &7.2    &6.9$\pm$0.1    &-0.3$\pm$0.1    &0.9857$\pm$0.0034   &97.4   &86.7$\pm$2.0   &-10.7$\pm$2.0   &0.9607$\pm$0.0109   \\
$W_V$           &33.6   &35.1$\pm$0.4   &+1.5$\pm$0.4    &0.9778$\pm$0.0017   &84.2   &75.1$\pm$0.9   &-9.1$\pm$0.9    &0.9721$\pm$0.0033   \\
$W_O$           &36.2   &29.3$\pm$0.1   &-6.9$\pm$0.1    &0.9978$\pm$0.0002   &83.5   &61.4$\pm$0.6   &-22.1$\pm$0.6   &0.9307$\pm$0.0039   \\
$W_{\text{in}}$ &48.1   &41.4$\pm$0.3   &-6.8$\pm$0.3    &0.9960$\pm$0.0003   &116.1  &83.1$\pm$0.8   &-33.0$\pm$0.8   &0.8566$\pm$0.0073   \\
$W_{\text{out}}$&33.7   &44.6$\pm$0.5   &+10.9$\pm$0.5   &0.9454$\pm$0.0015   &116.6  &100.4$\pm$0.4  &-16.2$\pm$0.4   &0.9572$\pm$0.0015   \\
Unembedding     &70.7   &62.0$\pm$0.6   &-8.7$\pm$0.6    &0.9451$\pm$0.0077   &73.3   &58.5$\pm$3.9   &-14.7$\pm$3.9   &0.9856$\pm$0.0082   \\
\bottomrule
\end{tabular}
\caption{Effective rank (ER) before and after noise injection, change in ER ($\Delta$ER), and cosine similarity (Cos) for each layer. 
Post-injection quantities are reported as mean $\pm$ std over 10 seeds; pre-injection ER is from the single pre-trained checkpoint.}
\label{tab:svs}
\end{table*}

The effective rank analysis reveals a pattern that varies across layer types. 
For the four attention projection matrices ($W_Q, W_K, W_V, W_O$) and the two MLP layers ($W_{\text{in}}, W_{\text{out}}$), the WLMD-equilibrium NN exhibits significantly higher ER than the AdamW-goldilocks NN prior to noise injection.
This indicates that the high-entropy state utilizes a substantially broader set of singular directions in these core layers, reflecting a richer feature representation of the learned task. 

After noise injection, both NNs experience a reduction in effective rank as the model adapts to accommodate the new noisy data. 
Notably, for the attention and MLP weight parameters, the absolute decrease in ER is larger for the WLMD-equilibrium NN than for the AdamW-goldilocks NN. 
Despite this greater reduction, the post-injection ER of the WLMD-equilibrium NN remains substantially higher than that of the AdamW-goldilocks NN across these layers. 
For the unembedding layer, the two NNs exhibit comparable ER prior to noise injection, but after injection, the WLMD-equilibrium NN shows a larger decrease, resulting in a post-injection ER lower than that of AdamW-goldilocks. 
These observations suggest that high-entropy NNs possess an abundance of representational capacity in their attention and MLP layers, which serves as a buffer: even after sacrificing a larger number of effective feature dimensions to accommodate new information, they still retain sufficient richness to preserve the original task knowledge. 

The cosine similarity between parameter vectors before and after noise-injection is higher for the AdamW-goldilocks NN compared to the WLMD-equilibrium NN, indicating that the WLMD-equilibrium NN undergoes substantially larger parameter displacement. 
This result aligns directly with the entropy perspective: because the WLMD-equilibrium NN corresponds to a high-entropy region of significantly larger volume in parameter space, its parameter point within this region remain in the same high-entropy basin even after considerable displacement, thereby preserving its generalizability on the original task. 
In contrast, the AdamW-goldilocks NN corresponds to a much narrower region, suffers from catastrophic forgetting when its parameters are displaced out of the low-entropy basin. 

\section{Related Work}

\subsection{Parameter-Space Geometry, Generalization, and Statistical Physics}

The relationship between the geometric structure of the parameter space and NN's generalization has been a central theme in deep learning theory. 
The flat minima hypothesis \cite{hochreiter1997flat} posits that flatter minima in the loss landscape correspond to better generalization, inspiring numerous theoretical and empirical investigations. 
However, subsequent work has revealed that sharp minima can also generalize well \cite{dinh2017sharp}, and that flatness measures based on the Hessian can be arbitrarily manipulated through simple rescaling the parameters without affecting its generalization performance. 

Statistical physics has provided a powerful lens for understanding these phenomena. 
Chaudhari et al. introduced entropy-SGD, linking the physics energy landscape to generalization through a local entropy objective that biases optimization toward wide valleys \cite{chaudhari2019entropy}. 
Bahri et al. provided a comprehensive review of statistical physics approaches to deep learning, covering topics from the loss landscape to the dynamics of optimization \cite{bahri2020statistical}. 

Within this tradition, the Boltzmann entropy of NN states has been introduced recently as a global measure of the volume in the parameter space corresponding to a given training loss and test performance \cite{yang2026high}. 
This entropy landscape framework offers a complementary perspective to local flatness measures, capturing the global structure of the solution space. 
Yang et al. established the high-entropy advantage in generalizability: NN states occupying larger parameter space volumes exhibit superior generalization compared to conventionally trained solutions, a finding validated across diverse tasks including arithmetic reasoning, tabular data, image recognition, and language modeling \cite{yang2026high}. 
Subsequent work by Zhang et al. further examined this advantage in the specific context of grokking on modular arithmetic tasks, framing the grokking phenomenon as a computational glass relaxation process and demonstrating that the high-entropy advantage is particularly significant in this setting \cite{zhang2026grokking}. 
The present study further extend this line of research by shifting the focus from generalization (measured by test metrics) to robustness (measured by knowledge retention under parameter space perturbations).

\subsection{Effective Rank: Measures, Generalization, and Robustness}
\label{subsec:ER}

The notion of effective rank provides a real-valued extension of matrix rank, quantifying the number of significant singular directions in a weight matrix \cite{roy2007effective}. 
For a weight matrix with singular values \(\sigma_1 \ge \sigma_2 \ge \dots \ge \sigma_k > 0\), the effective rank is defined as \(\text{ER}(W) = \exp(-\sum_i \tilde{\sigma}_i \log \tilde{\sigma}_i)\), where \(\tilde{\sigma}_i = \sigma_i / \sum_j \sigma_j\). 
This measure has been widely adopted as a tool for characterizing the intrinsic dimensionality of learned representations in NNs \cite{huh2021low}. 

A line of work has identified what is known as the \emph{low-rank simplicity bias} in deep NNs \cite{huh2021low}. 
Huh et al. demonstrated that deeper NNs are inductively biased to find solutions with lower effective rank embeddings, and conjectured that this bias exists because the volume of functions that map to low effective rank embeddings increases with depth \cite{huh2021low}. 
On natural data, these low-rank solutions are often the ones that generalize well. 
This bias has been observed to exist both at initialization and after training, and is resilient to hyper-parameters and learning methods \cite{huh2021low}.

The relationship between effective rank and robustness, however, is more nuanced and context-dependent. 
Langenberg et al. showed that adversarial training tends to promote simultaneously low-rank and sparse structure in weight matrices, as measured by effective rank and effective sparsity \cite{langenberg2019effect}. 
Furthermore, when low-rank structure is explicitly promoted via nuclear norm regularization, NNs show substantially improved adversarial robustness \cite{langenberg2019effect}. 
This indicates that in the context of adversarial training, low effective rank is associated with \emph{higher adversarial robustness}, which measures models' resistance to input-space perturbations with fixed parameters. 
This finding complements the low-rank simplicity bias observed in standard training. 
However, the underlying mechanisms differ: the low-rank solutions reached by standard gradient descent arise from the implicit regularization of the optimization dynamics, whereas the low-rank structure induced by adversarial training or explicit regularization is a consequence of carefully designed objectives that actively promote compression. 

Our work contributes a distinct perspective: high-entropy NNs sampled from the Boltzmann entropy landscape possess higher effective rank in attention and MLP layers, and this high effective rank is associated with stronger robustness to noise injections—complementing the existing literature where robustness has primarily been studied in the context of low-rank structures. 

\subsection{Catastrophic Forgetting and Robustness to Parameter-Space Perturbations}

Catastrophic forgetting is a phenomenon where NNs lose previously acquired knowledge upon learning new information. 
It has been a central challenge since its early identification \cite{mccloskey1989catastrophic}. 
Classical approaches to mitigating forgetting include elastic weight consolidation (EWC) \cite{kirkpatrick2017overcoming}, which penalizes changes to parameters important for previous tasks, and synaptic intelligence (SI) \cite{zenke2017continual}, which accumulates per-synaptic importance measures during training. 

Recent work has begun to explore the relationship between the geometric structure of parameter space and catastrophic forgetting. 
Brady Steele demonstrated that forgetting in low-rank adaptation (LoRA) is governed not simply by the rank of the adapter, but by the geometric relationship between task gradient subspaces \cite{steele2026subspace}. 
This finding suggests that the representational structure of the NN plays a fundamental role in determining its susceptibility to forgetting. 
Our noise injection experiment follows this line of inquiry by examining how the entropy of a NN state affects its ability to retain prior knowledge when learning new information.

\section{Conclusions and Discussion}
\subsection{Conclusions}

In this work, we have extended the high-entropy advantage beyond generalization to the domain of robustness. 
Using noise injection experiments on modular arithmetic tasks, we demonstrated that high-entropy NNs sampled from the Boltzmann entropy landscape maintain approximately 95\% test accuracy on the original task after memorizing new noisy data, whereas AdamW-trained NNs suffer from catastrophic forgetting, dropping to about 75\% under the same conditions. 
We term this hidden fragility behind apparent generalization the \textit{grokked illusion}. 
Through singular value decomposition of NN weights, we revealed that high-entropy NNs possess significantly higher effective rank in attention and MLP layers both before and after noise injection, indicating that richer feature representations serve as a buffer against parameter-space perturbations. 
These findings establish that perfect generalization does not imply equal robustness, and that the statistical physics framework can provide a principled lens for understanding both the quality and the resilience of learned representations. 

\subsection{Discussion}

Consistent with intuition, the robustness gap between the two NN types narrows as the injected noise becomes more structurally similar to the original task, suggesting that the degree of representational overlap between old and new information mediates the extent of forgetting. 

Our findings also speak to the broader literature on effective rank discussed. 
While previous work has shown that low-rank structures can arise from either implicit regularization (standard training) or explicit compression (adversarial training), we observe a third regime: high-entropy sampling yields high-rank representations that are intrinsically robust. 
The observation that high-entropy NNs exhibit both higher ER and stronger robustness suggests a possible connection between the two. 
High ER corresponds to a more uniform singular value spectrum, indicating that information is distributed across a broader set of feature dimensions. 
We speculate that this distributed representation provides a form of redundancy that allows the NN to absorb new information while retaining sufficient dimensions to preserve original task performance, which is consistent with the observation that WLMD-equilibrium NNs maintain higher ER after noise injection despite a larger decrease. 
Conceptually, this shares the same spirit of mixture-of-experts architectures, where distributing computation across multiple parameter subspaces can reduce reliance on any single computational route\cite{shazeer2017outrageously, li2025theory}. 
In our experiments, the representation redundancy is not explicitly enforced during conventional training: AdamW, even with strong regularization such as weight norm constraint, remains a loss-minimization procedure that does not bias toward high-entropy regions. 
WLMD, by contrast, explicitly biases the search toward states with larger parameter space volume under the same constraint. 
This difference in objective explains why the two types of NNs, despite identical weight norm and test accuracy, occupy fundamentally different regions of the parameter space and exhibit drastically different robustness behaviors. 

\subsection{Implications}

Our findings carry several implications beyond the specific setting of modular arithmetic. 
Firstly, the robustness advantage of high-entropy states offers a new perspective on catastrophic forgetting in continual learning. 
Our results suggest that the entropy of a NN's state could be a fundamental property associated with its robustness to forgetting. 
This opens the possibility of entropy-guided strategies for continual learning, where the goal is not merely to regularize updates but to actively position the NN in high-entropy regions that are intrinsically more robust. 

Secondly, the observed relationship between high effective rank and robustness may inform both the design and the evaluation of large language models (LLMs). 
Current LLM benchmarks face saturation and contamination issues, motivating a shift toward more challenging and dynamic evaluation settings, including interactive agent benchmarks and long-horizon reasoning tasks \cite{luo2025mcp, luo2026ultrahorizon, motwani2026longcot, zeng2026futurex}. 
Such settings expose LLMs' limitations bot captured by static task accuracy: their performance can degrade substantially as context length and task horizon increase \cite{du2025context}, while agents need to continually acquire new knowledge remain susceptible to catastrophic forgetting \cite{yang2026learning}.
The noise injection experiment we designed here provides a complementary dynamic evaluation principle: instead of evaluating what a trained model can perform on a fixed set of tasks, we can also measure whether its established capabilities remain intact when adapting to new tasks and information.
Moreover, this ability is also directly relevant to the post training and customization of LLMs, where the model is required to acquire new capabilities without degrading its pretrained ones. 
Therefore, the high-entropy robustness advantage we observe in the controlled modular arithmetic setting motivates further investigation on whether the broader spectral structure associated with high-effective-rank states can support more robust continual adaptation in foundation models, although this hypothesis remains to be tested at LLM scale.

Finally, our findings provide a different perspective on the relationship between effective rank and robustness.
Previous work has primarily examined this relationship in two contexts: standard training, where low-rank structures arise from implicit regularization, and adversarial training, where low-rank structures are actively promoted via explicit compression and confer robustness to input-space perturbations \cite{langenberg2019effect}. 
In contrast, we identify a third regime in which high-entropy sampling produces weight matrices with substantially broader singular spectra, and these states exhibit stronger retention under learning-induced parameter updates.
This suggests that the relationship between ER and robustness is not fixed and context dependent. 
Its functional consequences may depend on how the spectral structure is formed and how its dominant directions align with the learned computation.
In the high-entropy states studied here, a high ER may reflect a more distributed encoding that is less vulnerable to subsequent learning interference.

\subsection{Future Directions}
Several directions remain for future exploration. 
First, while modular arithmetic offers an ideal controlled setting for isolating the effect of entropy, extending these findings to more complex tasks (such as natural language processing or image classification) and to deeper or more diverse architectures (particularly LLMs) is a natural next step. 
Verifying whether the high-entropy robustness advantage persists in different situations and how it scales with model size would strengthen the generality of our conclusions. 

Second, although effective rank provides compelling correlational evidence linking high entropy to robustness, establishing the causal relationship between entropy, effective rank, and robustness merits further investigation. 
Interventions that directly manipulate effective rank during training, or analytical studies of the entropy landscape's geometry, could help clarify whether high effective rank is a cause or merely a correlate of robustness. 

Third, the intrinsic robustness of high-entropy states suggests potential algorithmic developments. 
If high-entropy states are inherently more robust, future work could design training procedures that explicitly bias optimization toward high-entropy regions while maintaining generalization performance. 
One attempt is the WanD optimizer which employs 1D WLMD to search for high entropy solutions, though it faces the issue of training instability \cite{zhang2026grokking}. 
Developing such entropy-guided training methods and testing them on standard continual learning benchmarks would translate our fundamental findings into practical methods for reducing catastrophic forgetting. 

\section*{Ethics Statement}

This work involves only synthetic data generated from modular arithmetic operations and does not involve any human subjects, personal data, or sensitive information. 
All experiments were conducted on standard benchmarks and synthetic datasets that do not raise privacy, fairness, or safety concerns. 
The noise injection experiments in this study are designed to evaluate NN's robustness and do not involve any real-world applications that could cause harm. 
We do not foresee any direct negative societal impacts of this work, as it is a fundamental study of NN properties rather than a deployed system. 
All code and models used in this paper will be made publicly available upon publication to facilitate reproducibility. 
No large language models were used in the writing or generation of this manuscript beyond editing and polishing. 
All experimental design, data collection, analysis, and interpretation of results were performed entirely by the human authors. 
The authors assume full responsibility for the content of this paper. 

\section{Acknowledgments}
The authors thank National Natural Science Foundation of China for supporting this research (Grant 12405043, G.Z.).
We also thank computational resources provided by Bridges-2 at Pittsburgh Supercomputing Center and DeltaAI at National Center for Supercomputing Applications through ACCESS allocation CIS230096 (E.Y.).

\bibliography{aaai2027}


\end{document}